\documentclass[runningheads]{llncs}
\usepackage[T1]{fontenc}
\usepackage{graphicx}
\usepackage{color}
\usepackage{subfigure}
\usepackage{tikz}
\usepackage{multirow}
\usepackage[export]{adjustbox}

\usepackage{hyperref}
\hypersetup{
    colorlinks=true,
    linkcolor=blue,
    filecolor=magenta,      
    urlcolor=cyan,
    }

\begin{document}

\title{S-SYNTH: Knowledge-Based, Synthetic Generation of Skin Images}
\author{Andrea Kim\inst{*} \and
Niloufar Saharkhiz\inst{*} \and
Elena Sizikova\inst{*} \and \\
Miguel Lago \and
Berkman Sahiner \and
Jana Delfino \and
Aldo Badano}
\authorrunning{A. Kim et al.}
\institute{Office of Science and Engineering Laboratories\\ 
Center for Devices and Radiological Health\\ 
U.S. Food and Drug Administration\\ 
Silver Spring, MD 20993 USA \\
\email{elena.sizikova@fda.hhs.gov} \\
\inst{*}-These authors contributed equally to this work.}

\maketitle              
\begin{abstract}
Development of artificial intelligence (AI) techniques in medical imaging requires access to large-scale and diverse datasets for training and evaluation. In dermatology, obtaining such datasets remains challenging due to significant variations in patient populations, illumination conditions, and acquisition system characteristics. In this work, we propose S-SYNTH, the first knowledge-based, adaptable open-source skin simulation framework to rapidly generate synthetic skin, 3D models and digitally rendered images, using an anatomically inspired multi-layer, multi-component skin and growing lesion model. The skin model allows for controlled variation in skin appearance, such as skin color, presence of hair, lesion shape, and blood fraction among other parameters. We use this framework to study the effect of possible variations on the development and evaluation of AI models for skin lesion segmentation, and show that results obtained using synthetic data follow similar comparative trends as real dermatologic images, while mitigating biases and limitations from existing datasets including small dataset size, lack of diversity, and underrepresentation.

\end{abstract}

\section{Introduction}
Robust and generalizable artificial intelligence (AI) for medical applications requires large datasets representative of intended populations and their subgroups, relying on a time-consuming and laborious annotation process by clinical experts (e.g. labeling pixel-level segmentation masks or identifying findings). In this work, we explore the potential of a skin simulation model and the practical utility of resulting synthetic images in development and evaluation of AI-based  techniques. 

Segmenting and labeling dermatologic images is time-consuming and challenging, and thus public dermatologic datasets generally contain few examples and may not equitably represent intended patient populations~\cite{chen2023evaluation,groger2023towards}. For instance, the majority of the publicly available dermatologic images are associated with lighter skin populations, and dark skin tones are typically under-represented~\cite{kinyanjui2019estimating}. This class imbalance has shown to negatively affect darker skin tones in skin lesion segmentation tasks \cite{benvcevic2024understanding}. Such challenges may introduce biases in training and evaluation of medical AI devices for dermatologic applications. 

To address these limitations, we propose to simulate dermatologic images with a knowledge-based approach~\cite{badano2023stochastic}, where properties of skin and lesions, determined via detailed physics models, along with rendering conditions are systematically varied. We created S-SYNTH, an open-source skin simulation framework for generating synthetic dermatologic images, and show that subgroup analysis using S-SYNTH images closely mimics comparative performance trends in real-patient skin datasets, while mitigating effects of mislabeled examples and limitations of a limited patient evaluation dataset. Our contributions can be summarized as follows:
\begin{itemize}
\item We describe S-SYNTH, an open-source, flexible framework for creation of highly-detailed 3D skin models and digitally rendered synthetic images of diverse human skin tones, with full control of underlying parameters and the image formation process\footnote{Code and supporting data are available at: \url{https://github.com/DIDSR/ssynth-release}}. 
\item We systematically evaluate S-SYNTH synthetic images for training and testing applications. Specifically, we show S-SYNTH synthetic images improve segmentation performance when only a limited set of real images is available for training. We also show comparative trends between S-SYNTH synthetic images and real-patient examples (according to skin color and lesion size) are similar.
\end{itemize}

\section{Related Work}    
A particular challenge in practical development and evaluation of dermatologic AI is the lack of labeled datasets. Of the few public labeled datasets that exist, only a fraction have fine grained annotations, such as pixel-level segmentation levels for lesions~\cite{mirikharaji2023survey} or hair~\cite{hossain2023skin}. The annotation process is error-prone and can have large variations across truthers (particularly when datasets are collected for different purposes and no consistent labeling protocol is established). In part due to these annotation difficulties, there is a lack of transparency and potential bias within the datasets used for dermatologic AI studies. A recent investigation found that from a total of 70 unique studies that involved AI models for dermoscopic imaging, only about 7 (10\%) included skin tone information for at least one dataset used~\cite{daneshjou2021lack}. Synthetically generated data has been explored to address these challenges. Generative models have been used to create synthetic skin lesions to tackle the class imbalance problem and improve the performance of the skin lesion classifiers~\cite{baur2018melanogans,behara2023skin,ghorbani1911dermgan} or enhance lesion segmentation~\cite{chi2018controlled,oliveira2020controllable}. Rezk et al. have shown that synthetic skin images generated using style transfer and deep blending improve skin lesion classification accuracy by increasing skin color diversity \cite{rezk2022improving}. More recently, diffusion models have been used to generate synthetic images to improve skin disease classifier performance \cite{sagers2022improving,sagers2023augmenting}. Nevertheless, use of AI to generate synthetic images is limited by the requirement of constant supervision and adjustment of hyperparameters~\cite{baur2018melanogans,behara2023skin}, presence of artifacts, such as skin hair or lighting effects, within the original training dataset~\cite{oliveira2020controllable}, and a lack of ability to represent all physical and morphological features of disease \cite{rezk2022improving} and image acquisition systems (see \cite{mirikharaji2023survey} for more information). 

\begin{figure}[tb]
 \centering
  \includegraphics[width=\columnwidth]{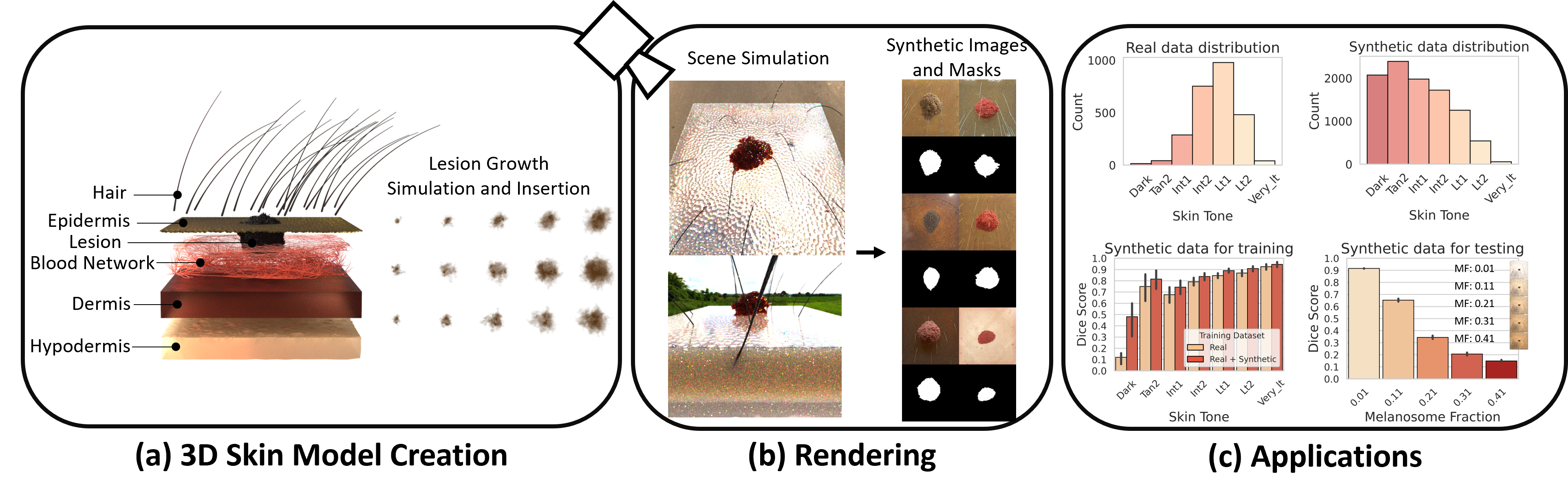}
 \caption{Overview of (a) a digital skin model generated in Houdini, and an example projection of 3 different skin lesion volumes at 5 growing time points, (b) digital rendering, with examples of generated synthetic images and their corresponding lesion masks, and (c) distribution of skin tone for real-patient images (ISIC) and synthetic images, as well as application of synthetic images for training and testing of an AI device for a skin lesion segmentation task.}
 \label{fig:skin_model}
\end{figure}

\section{S-SYNTH: Skin Simulation Framework}
Our approach for generating synthetic skin involves the construction of a 3D digital object model comprising of skin tissue (epidermis, dermis, hypodermis), blood network, hair, and a lesion, as shown in Fig.~\ref{fig:skin_model}a. This process is implemented and automated in Houdini~\cite{houdini}, a software for 3D modeling, animation, and visual effects, via a Python API. Once created, each model is subsequently processed through a research-oriented rendering system, Mitsuba 3~\cite{jakob2022mitsuba3}, to generate each synthetic rendering (Fig~\ref{fig:skin_model}b). The approach can be automated to create large databases with controlled variation for various AI analysis applications ((Fig~\ref{fig:skin_model}c)). In the following sections, we describe the digital skin model and rendering process.

\subsection{Digital Skin Model}
\subsubsection{Skin Tissue} Each skin sample is created from a multi-layer model, with each layer representing a component of skin tissue. Thicknesses of the topmost two layers are based on values reported in \cite{walters2002structure}: epidermis (20-150 $\mu m$) and dermis (1-4 mm). To introduce geometric variability, we incorporated surface roughness (noise) into the top surfaces of epidermis, hypodermis, and papillary dermis (located between the epidermis and dermis). Roughness values were randomized within predefined constraints, enhancing the naturalistic appearance and diversity of the generated skin models.

\subsubsection{Blood Network and Hair}
The blood network model is created by solving a shortest path problem on randomly distributed points within a tetrahedral mesh generated from a primitive cube. Points from the bottom of the cube, corresponding to the lower blood network capillaries, were designated as start points, while those from top of the cube, representing the upper capillaries, were designated as end points. To generate each blood network model, the numbers and positions of starting and ending points were randomized. Hair models were constructed by manipulating hair properties (density, length, distribution, thickness, and curvature) in Houdini to generate a diverse range of hair apprearances.

\subsubsection{Growing Lesion Model}
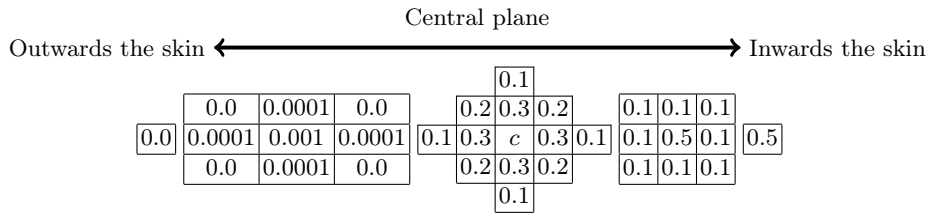
\begin{figure}[]
    \noindent\centering
    \begin{tikzpicture}
        \node[left=7cm](out) {Outwards the skin};
        \node[right=0cm](in) {Inwards the skin};
        \draw[<->, line width=1.5pt] (out.east) -- node[above=1mm, align=center] {Central plane} (in.west);
    \end{tikzpicture}
    \begin{tabular}{|c|}
        \hline
        0.0\\
        \hline
    \end{tabular}
    \begin{tabular}{|c|c|c|}
        \hline
         0.0&0.0001&0.0\\ \hline
         0.0001&0.001&0.0001\\ \hline
         0.0&0.0001&0.0\\ \hline
    \end{tabular}    
    \begin{tabular}{cc|c|cc}
        \cline{3-3}
         \multicolumn{2}{c|}{ } & 0.1 & \multicolumn{2}{c}{ } \\\cline{2-4}
         \multicolumn{1}{c|}{ } & 0.2&0.3&0.2&\multicolumn{1}{|c}{ }\\\cline{1-5}
         \multicolumn{1}{|c|}{0.1}& 0.3&$c$&0.3&\multicolumn{1}{|c|}{0.1} \\\cline{1-5}
         \multicolumn{1}{c|}{ } & 0.2&0.3&0.2&\multicolumn{1}{|c}{ }\\\cline{2-4}
         & & 0.1 & & \\\cline{3-3}
    \end{tabular}
    \begin{tabular}{|c|c|c|}
        \hline
         0.1&0.1&0.1\\ \hline
         0.1&0.5&0.1\\ \hline
         0.1&0.1&0.1\\ \hline
    \end{tabular}
    \begin{tabular}{|c|}
        \hline
        0.5\\
        \hline
    \end{tabular} 
    \caption{Growing direction probabilities for each cell $c$ in the 3D plane. Note that values are zeros in the \textit{outwards} direction, to avoid the lesion growth outside the skin.}
    \label{tab:growing_directions}
\end{figure}

We developed a probabilistic, growing lesion model that generates a 3D volume with stochastic growth of lesion shape and size, based on \cite{sengupta2021computational}. Growth starts with one active cell (i.e., voxel). We define \textit{active cells} as those cells that exist and can grow at a given time point. Within each time point, we iterated over the active cells $c$ and selected them for growth based on probabilities (see Fig. ~\ref{tab:growing_directions}). To control the growing direction, the probability of an active cell to grow outside the skin is set to extremely low, while the probability for inwards growing is set to be high. Probabilities were based on a Gaussian distribution $G$, and are updated in each time point for added variability.

To control for lesion shape irregularity, we set an irregularity probability for each cell $C_p$, which starts $C_i$ number of recursive iterations of the same growing algorithm on that cell's location, independent of the regular growth. Irregular cells can themselves trigger another recursive growth up to a maximum of $C_r$ recursions. We generated both regular  ($C_p=0.0001$) and irregular lesions ($C_p=0.001$).  Sample images of lesions and growth evolution can be seen in Fig ~\ref{fig:skin_model}.  Additional details can be found in the Supplementary Material. Once created, the 3D lesion volume is inserted into the skin tissue model (see \cite{kim2023automated} for more details).

\subsection{Rendering and Image Formation}

\subsubsection*{Optical Material Properties}
Melanosomes are the primary contributors to optical absorption in the epidermis. In the dermis, optical absorption is mainly attributed to blood, while, in the hypodermis, lipid and fat components are the predominant sources of optical absorption. Thus, each skin layer is assigned an  optical material, containing the index of refraction (IOR) and the spectral distribution of the absorption and scattering coefficients.  Following the methodology established by Jacques et al.~\cite{jacques2013optical}, we calculated the optical absorption properties for each the four distinct tissue types: epidermis, dermis, hypodermis, and blood. We model reduced scattering of tissue using $\mu'_{s} = a(\lambda/500)^{-b}$, where $\lambda$ is the light wavelength \cite{bench2020toward}. Finally, for each skin layer material, we define a surface scattering model using a bidirectional scattering distribution function (BSDF), based on \cite{walter2007microfacet}.

\subsubsection{Lighting and Camera} 
\begin{figure}[htb]
 \centering
\includegraphics[width=\columnwidth]{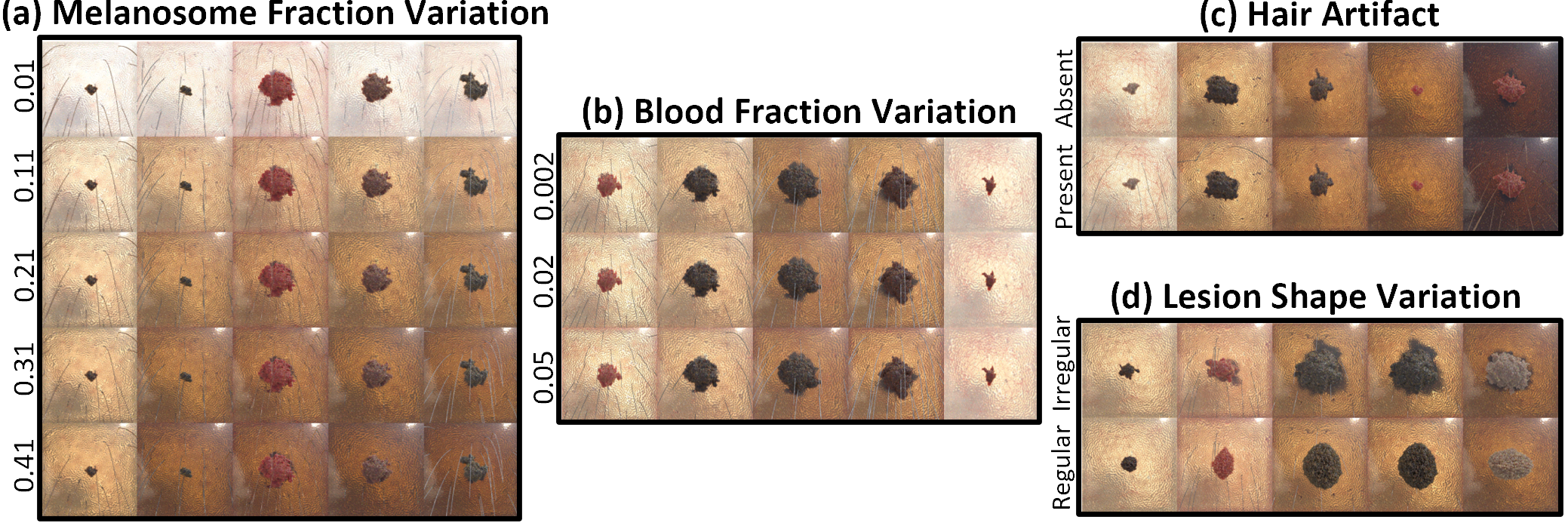}
 \caption{Examples of S-SYNTH images generated with variations of (a) melanosome fraction, (b) blood fraction, (c) hair artifact, and (d) lesion shape.} 
 \label{fig:examples}
\end{figure}

To account for lighting variations, we render each skin model using a collection of High Dynamic Range Imaging (HDRI) images capturing various  environmental lighting conditions from \cite{polyhaven}. The Volumetric Path Tracer is a rendering algorithm that simulates the paths taken by light as it interacts with the 3D scene. It is particularly effective for scenes with volumetric effects or materials with complex light interactions. Spectral multiple importance sampling is a technique used to address the spectral nature of light. In scenes with materials that exhibit variations in spectral properties (e.g., spectral absorption), this method accounts for the complex interactions of light with materials across different wavelengths, such as those encountered in biological tissues. A perspective sensor with Mitsuba's HDR Film is placed 1.5cm above the skin with a field of view (FOV) of 75 degrees, facing perpendicularly to the top surface of the epidermis. The Integrator (Mitsuba 3's plugin for the rendering technique) is set to Volumetric Path Tracer with spectral multiple importance sampling. We chose this specific configuration due to its ability to simulate spectrally-varying optical properties across varying skin tones in simulated lighting conditions. We use 124 samples per pixel (SPP) and render images at resolution 1024x1024 pixels. Each image and mask take about three minutes to generate on a GPU. Sample generated images can be seen in Fig.~\ref{fig:examples}. 
 
\subsection{AI Device Description}
To demonstrate the usage of S-SYNTH  for augmenting real patient data, we experiment with real and synthetic examples on the task of skin lesion segmentation. which is important for timely treatment decisions~\cite{ruan2023ege}. We rely on DermoSegDiff~\cite{bozorgpour2023dermosegdiff}, a state of the art diffusion-based skin segmentation model, to differentiate lesions from the background region. We used the ``dsd\_i01'' configuration, with \verb|dim_x=16|, \verb|dim_g=8|, and default training parameters. Images were segmented at resolution 128x128, and results were evaluated with Dice coefficient. We trained one model per experiment and reported standard deviation over test images.

\subsubsection*{Datasets} For real-patient datasets, we used two publicly available skin lesion segmentation datasets: ISIC18~\cite{codella2019skin} (ISIC) and HAM10K~\cite{tschandl2018ham10000} (HAM). The same pre-processing and data splits described in \cite{codella2019skin,azad2022transnorm} were applied, where the ISIC and HAM images were divided into (7200, 1800, 1015) and (1815, 259, 520) for the baseline model (training, validation, test), respectively. We used the individual typology angle (ITA) metric~\cite{kinyanjui2019estimating} to estimate skin tone from the non-lesion area. Finally, to better match lesion sizes across real and synthetic examples, S-SYNTH images were cropped to ensure a variety of lesion sizes. With S-SYNTH, we generated 10,000 images with randomized model properties and lighting, and subsequently 19,965 testing images with controlled variation: 5,445 with blood variation, 9,075 with melanosome fraction variation, 3,630 with different lesion regularity, and 1,815 without hair.   
\section{Results and Discussion}    
We present segmentation performance based on three aspects: impact of training set composition comprising different ratios of real and synthetic images, impact of various physiological or rendering parameters of the synthetic images when used as test set, and impact of specific characteristics of synthetic images that are measurable on real images.

\subsubsection*{Synthetic Data for Training} \label{sec:res_training}
\begin{figure}[t]
    \centering
    \subfigure[Real data percentage]{\includegraphics[width=0.32\columnwidth]{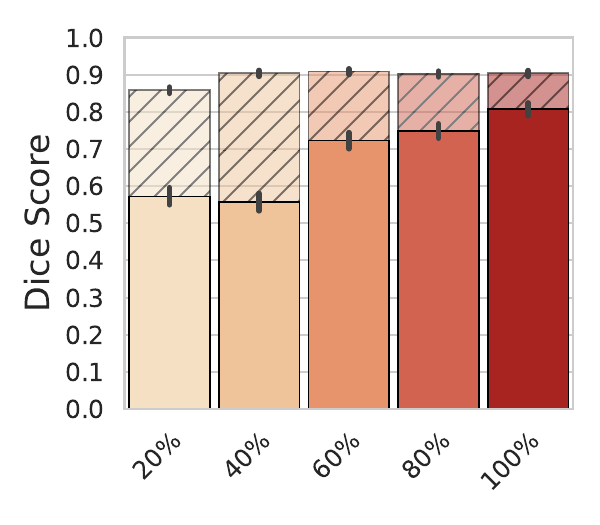}}
    \subfigure[\centering {Ratio of real to synthetic}]{\includegraphics[width=0.32\columnwidth]{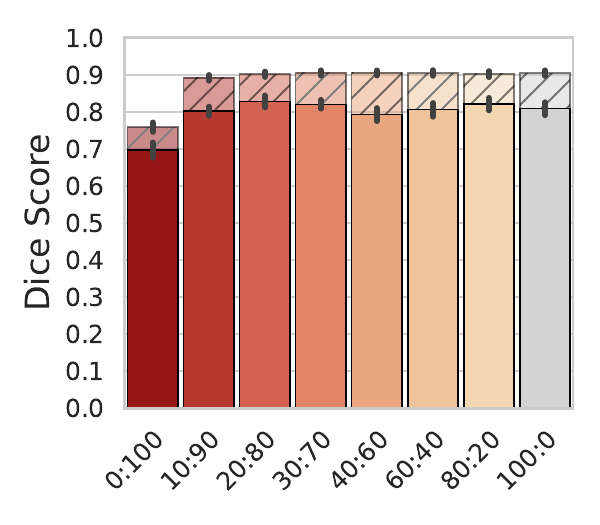}}
    \hfill
    \subfigure[Ratio of real to synthetic]{\includegraphics[width=0.32\columnwidth]{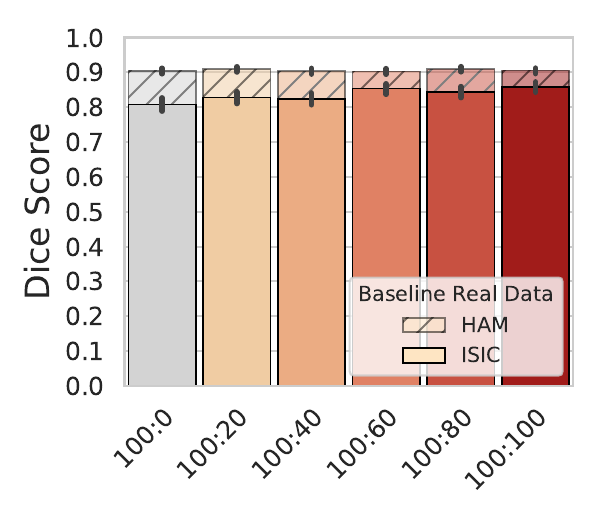}}
    \caption{Model performance changes when the training data is composed of (a) different numbers of the real images, (b) different proportions of real images replaced with synthetic images (c) different proportions of synthetic images added to real images.}
    \label{fig:data_subsets}

\end{figure}

We systematically evaluated the effect of the training data composition on the real test set performance (similar to the popular Train-Synthetic-Test-Real (TSTR) protocol~\cite{esteban2017real}) on both ISIC and HAM. As shown in Fig.~\ref{fig:data_subsets}a, performance improved when more real examples are available for training. When real images are replaced with synthetic images, the resulting performance is comparable to that of the baseline model trained on full ISIC (or HAM) training set (Fig.~\ref{fig:data_subsets}b). However, performance dropped when the training set was fully composed of synthetic images, presumably due to the domain shift between the real and synthetic images~\cite{guan2021domain}. More importantly, each model that was trained on a particular subset of the ISIC (or HAM) and supplemented with synthetic images resulted in a higher Dice score than the model that was trained on the same subset of only the real images. Finally, when full training sets were supplemented with synthetic images (Fig.~\ref{fig:data_subsets}c), performance improved for ISIC, but saturated for HAM, likely because ISIC was too small to learn all modes of variation, which synthetic data helped address. We also found that performance for darker skin tones improved with the addition of synthetic examples (see Supplementary Material). 

\subsubsection*{Synthetic Data for Testing} \label{sec:synth_test}
\begin{figure}[htb]
    \centering
    \subfigure[BF Variation]{\includegraphics[width=0.24\columnwidth]{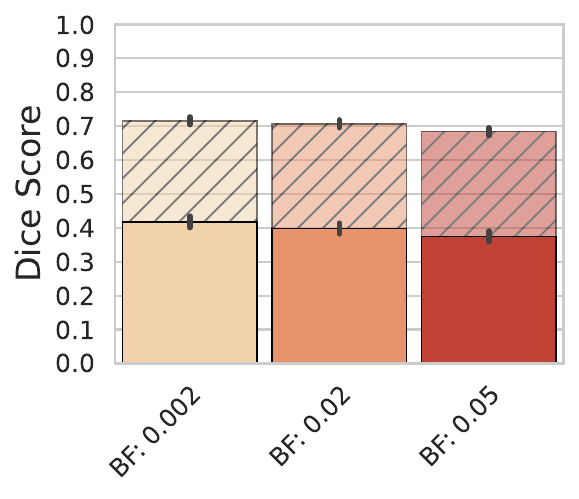}} 
    \subfigure[MF variation]{\includegraphics[width=0.24\columnwidth]{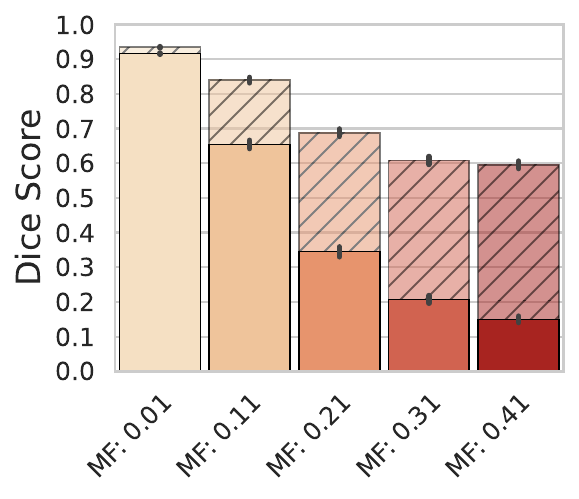}}
    \subfigure[Lesion Shape]{\includegraphics[width=0.24\columnwidth]{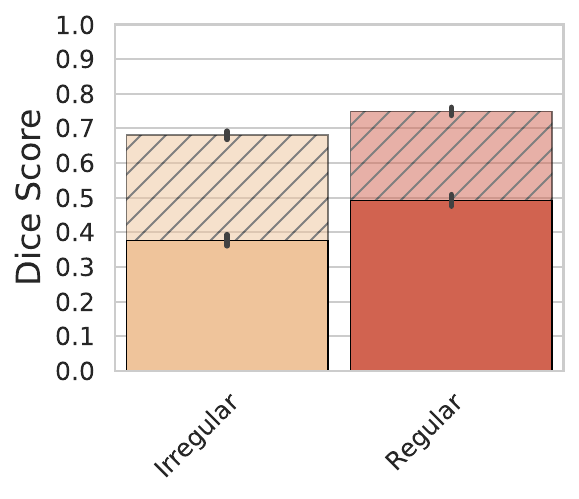}} 
    \subfigure[Hair Artifact]{\includegraphics[width=0.24\columnwidth]{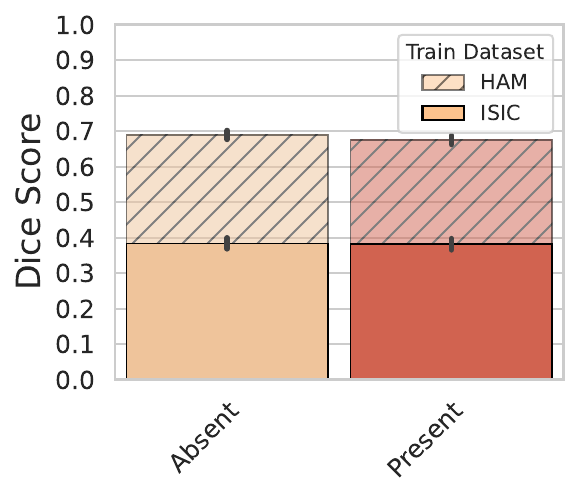}}
    \caption{Model performance when trained on real-patient datasets (ISIC, HAM) and tested on synthetic (S-SYNTH) images generated with different parameters. BF: blood fraction, MF: melanosome fraction.} \label{fig:test_agg}
\end{figure}

To check whether synthetic images can produce meaningful performance trends, we evaluated models trained on real examples using synthetic test images. Performance dropped for increasing blood fraction (Fig. \ref{fig:test_agg}a), and increasing melanosome fraction (Fig. \ref{fig:test_agg}b) of the S-SYNTH images. Higher melanosome fraction corresponds to darker skin, and the corresponding performance drop is a known bias in dermatologic AI~\cite{rezk2022improving}. Models tested on lesions with regular shape exhibited higher performance compared to the ones tested on irregular lesions (Fig. \ref{fig:test_agg}c). Similar to \cite{li2021digital}, we found that presence of hair slightly decreased performance (Fig. \ref{fig:test_agg}d). 

\begin{figure}[htb]
    \centering
    \subfigure[Skin Tone]{\includegraphics[width=0.32\columnwidth]{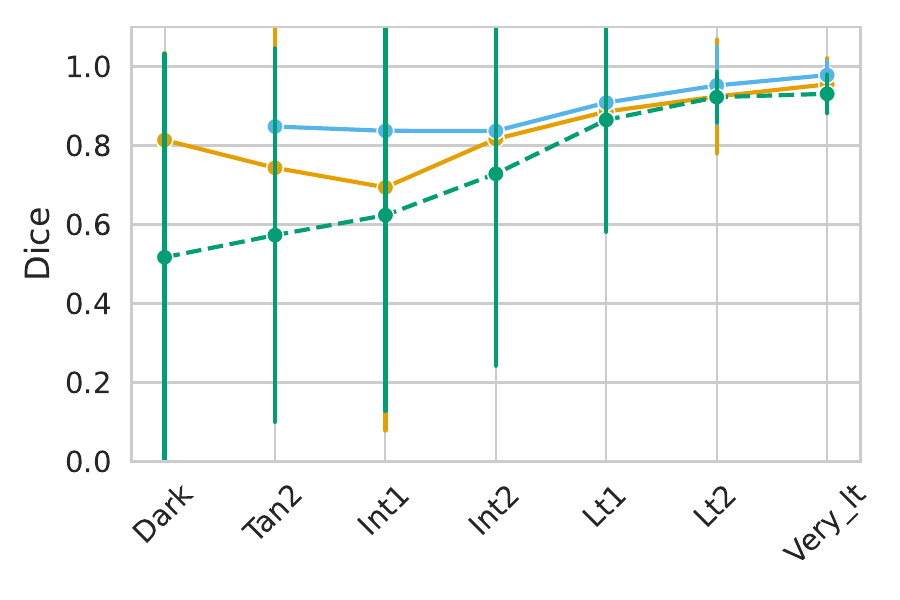}} 
    \subfigure[Lesion Circularity]{\includegraphics[width=0.32\columnwidth]{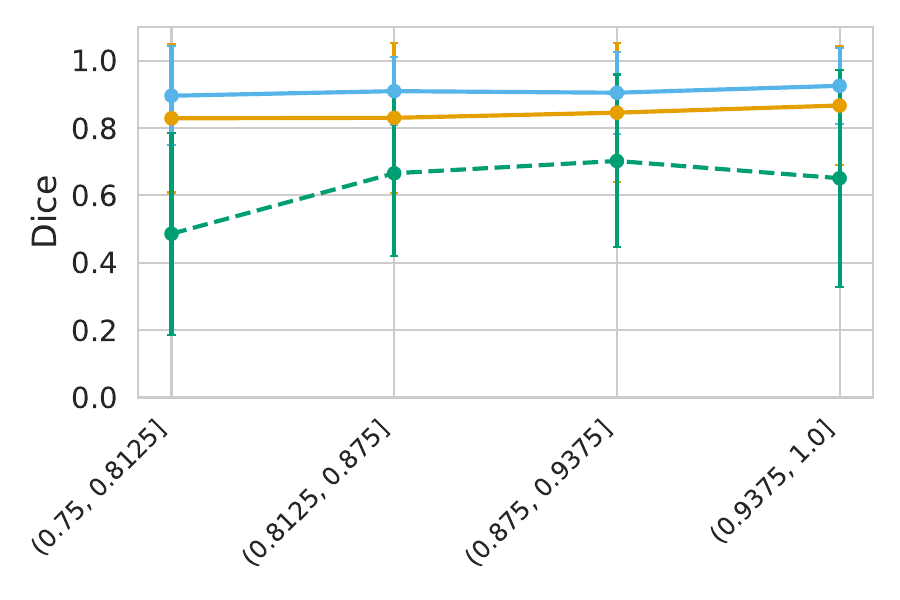}} 
    \hfill
    \subfigure[Relative Lesion Area]{\includegraphics[width=0.32\columnwidth]{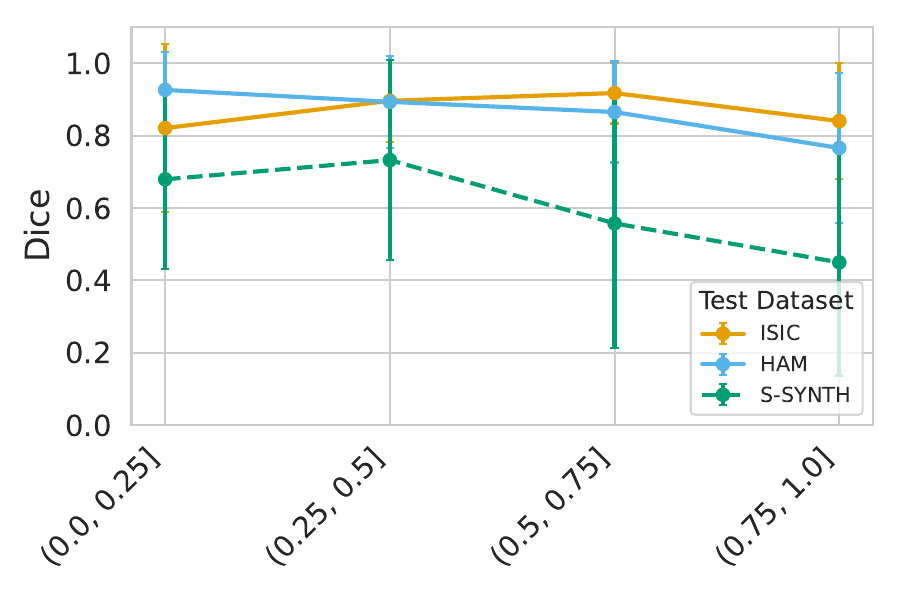}}
    \caption{Performance as a function of: (a) skin tone, (b) lesion circularity, and (c) relative lesion area for models trained on HAM dataset and tested on real (HAM, ISIC) or synthetic (S-SYNTH) images.} 
    \label{fig:comparative_analysis}
\end{figure}

To assess the similarity between the real and synthetic images in measurable characteristics, we evaluated performance changes as a function of skin color (ITA), lesion circularity, and lesion relative area. Although performance on synthetic images was lower compared to real images, similar trends were observed. Specifically, performance in both synthetic and real examples increased with lighter skin (Fig. \ref{fig:comparative_analysis}a). As illustrated in (Fig. \ref{fig:comparative_analysis}b), the impact of lesion circularity on model performance showed similar trends within the circularity range shared between the real and synthetic images, where more circular images were easier to segment. Finally, performance dropped with larger relative lesion areas (Fig. \ref{fig:comparative_analysis}c), possibly due to less precise delineations in larger lesions in the training patient data. 

\subsubsection{Limitations}
S-SYNTH is the first skin simulation framework that generates synthetic skin images with controllable variations. There are a number of limitations to our work. First, S-SYNTH does not model any specific disease presentation, and additional  work needs to be performed to evaluate the realism of the generated examples. Second, the current rendering techniques utilizes an RGB camera setup, and additional modifications allowing for multispectral imaging, when validated, might be useful for more advanced algorithms that use a broader variety of wavelengths. 

\section{Conclusion and Future Work}    
We presented S-SYNTH, a novel, knowledge-based simulation pipeline for synthetic generation of dermoscopic images and skin lesions. S-SYNTH procedurally generates multi-layer 3D skin models, with consideration to optical skin characteristics, and digitally renders synthetic images under realistic lighting conditions. Using S-SYNTH, we generated realistic and varied skin surface models for sub-surface scattering simulation in the context of skin lesion segmentation, and compared the resulting synthetic images to public dermatologic benchmarks. We showed that examples created using S-SYNTH can be used to augment limited real datasets and identify performance trends.

\bibliographystyle{unsrturl}
\bibliography{main}
\clearpage

%%%%%%%%%%%%%%%%%%%%%%%%%%%%%%%%%%%%%%%%%%%%%%%%%%%%%%%%%%%%%%%%%%%%%%%%%%%%%%%%%%%%%%%%%%%%%%%%%%%%%%%%%%%%%%%%%%%%%%%%%%%%%%%%%%%%%%%%%%%%%%%%%%%%%%%%%%%%%%%%%%%%%%%%%%%%%%%%%%%%%%%%%%%%%%%%%%%%%%%%%%%%%%%%%%%%%%%%%%%%%%%%%%%%%%%%%%%%%%%%%%%%%%%%%%%%%%%%%%%%%%%%%%%%%%%%%%%%%%%%%%%%%%%%%%%%%%%%%%%%%%%%%%%%%%%%%%%%%%%%%%%%%%%%%%%%%%%%%%%%%%%%%%%%%%%%%%%%%%%%%%%%%%%%%%%%%%%%%%%%%%%%%%%%%%%%%%%%%%%%%%%%%%%%%%%%%%%%%%%%%%%%%%%%%%%%%%%%%%%%%%%%%%%%%%%%%%%%%%%%%%%%%%%%%%%%%%%%%%%%%%%%%%%%%%%%%%%%%%%%%%%%%%%%%%%%%%%%%%%%%%%%%%%%%%%%%%%%%%%%%%%%%%%%%%%%%%%%%%%%%%%%%%%%%%%%%%%%%%%%%%%%%%%%%%%%%%%%%%%%%%%%%%%%%%%%%%%%%%%%%%%%%%%%%%%%%%%%%%%%%%%%%%%%%%%%%%%%%%%%%%%%%%%%%%%%%%%%%%%%%%%%%%%%%%%%%%%%%%%%%%%%%%%%%%%%%%%%%%%%%%%%%%%%%%%%%%%%%%%%%%%%%%%%%%%%%%%%%%%%%%%%%%%%%%%%%%%%%%%%%%%%%%%%%%%%%%%%%%%%%%%%%%%%%%%%%%%%%%%%%%%%%%%%%%%%%%%%%%%%%%%%%%%%%%%%%%%%%%%%%%%%%%%%%%%%%%%%%%%%%%%%%%%%%%%%%%%%%%%%%%%%%%%%%%%%%%%%%%%%%%%%%%%%%%%%%%%%%%%%%%%%%%%%%%%%%%%%%%%%%%%%%%%%%%%%%%%%%%%%%%%%%%%%%%%%%%%%%%%%%%%%%%%%%%%%%%%%%%%%%%%%%%%%%%%%%%%%%%%%%%%%%%%%%%%%%%%%%%%%%%%%%%%%%%%%%%%%%%%%%%%%%%%%%%%%%%%%%%%%%%%%%%%

\section{Supplementary Materials}

 \begin{figure}[htb]
 \centering
 \subfigure[HAM dataset]{\includegraphics[width=0.32\columnwidth]{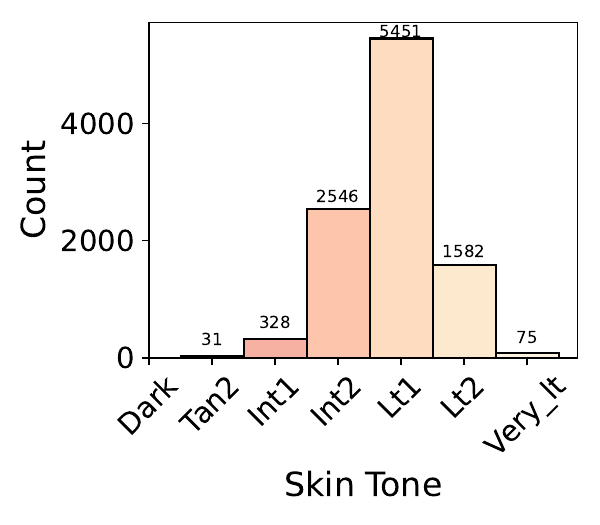}}
 \subfigure[ISIC dataset]{\includegraphics[width=0.32\columnwidth]{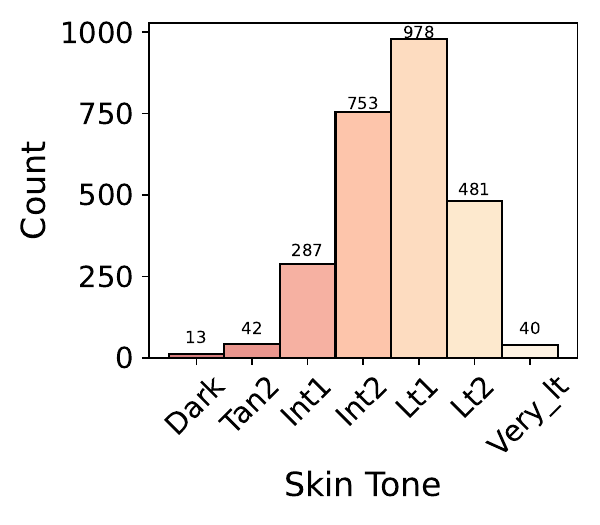}}
  \subfigure[S-SYNTH dataset]{\includegraphics[width=0.32\columnwidth]{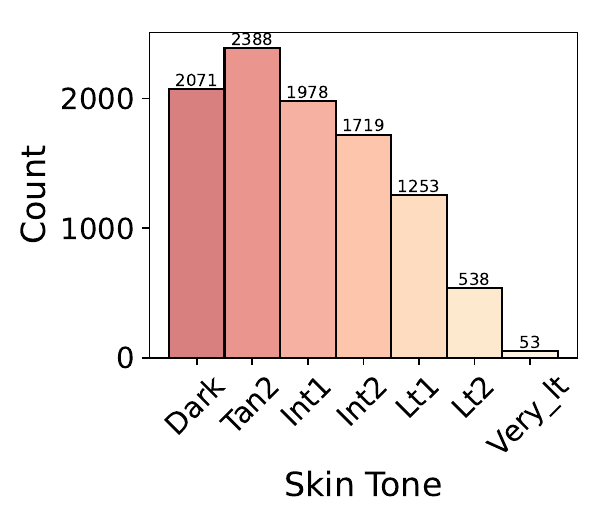}}
 \caption{ Distribution of skin tone category based on the mean ITA score of the non-diseased region for (a) HAM dataset with 10015 images, (b) ISIC dataset with 2594 images, and (c) S-SYNTH dataset with 10000 images. The categories are estimated based on \cite{kinyanjui2019estimating}, however, the examples categorized as "dark" and "tan1" are combined due to the limited number of examples in each group for the real-patient datasets.} 
 \label{fig:skin_tone_hist}
\end{figure}

\begin{table}[htb]
    \centering
    \begin{tabular}{|c|c|}
    \hline
        Growing step & $S\sim U(1,2)$\\
        Probabilities update & $G(\mu=0, \sigma= 0.5)$ \\
        Cancer probability & $C_p$ = 0.0001\\
        Cancer iterations & $C_i = 10$\\
        Maximum cancer recursion & $C_r$ = 2 \\ \hline
    \end{tabular}
    \caption{Assigned values and probabilities to the skin growing lesion parameters.}
    \label{tab:skin-lesion-probabilites}
\end{table}

\begin{figure}[h!]
 \centering
\includegraphics[width=\columnwidth]{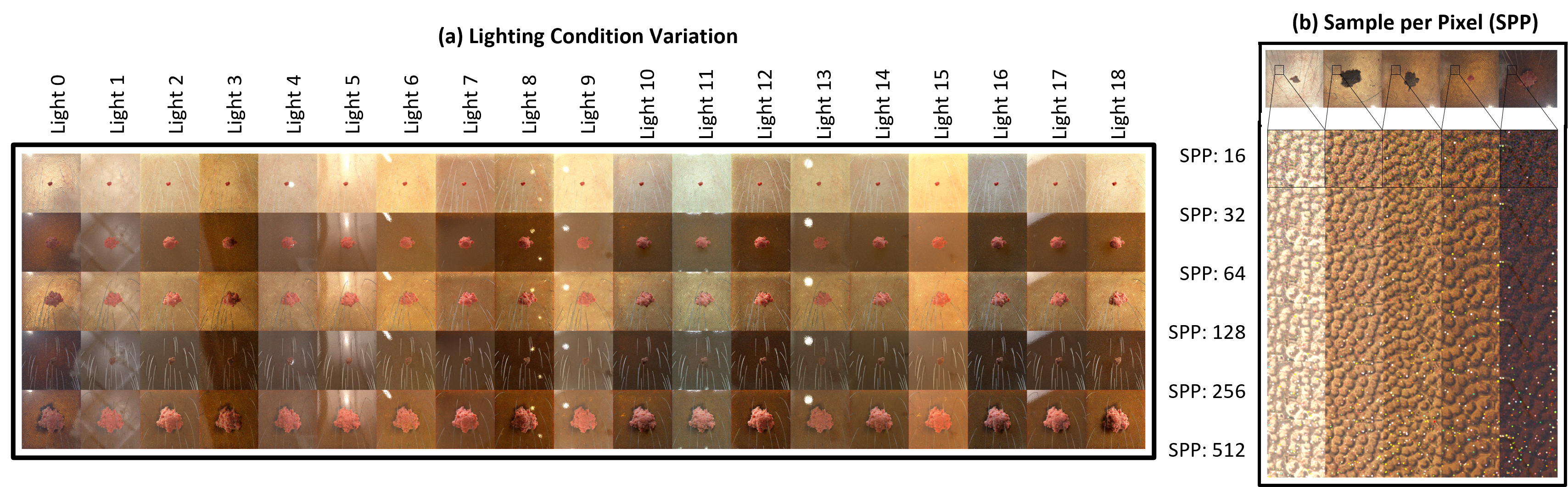}
 \caption{Examples of S-SYNTH images generated with variations of (a) lighting conditions, (b) samples per pixel (SPP).} 
 \label{fig:examples_light_SPP}
\end{figure}

\begin{figure}[htb]
    \centering
    \subfigure[Examples of uncropped (top row) and cropped S-SYNTH images (bottom row)]{\includegraphics[width=0.56\columnwidth]{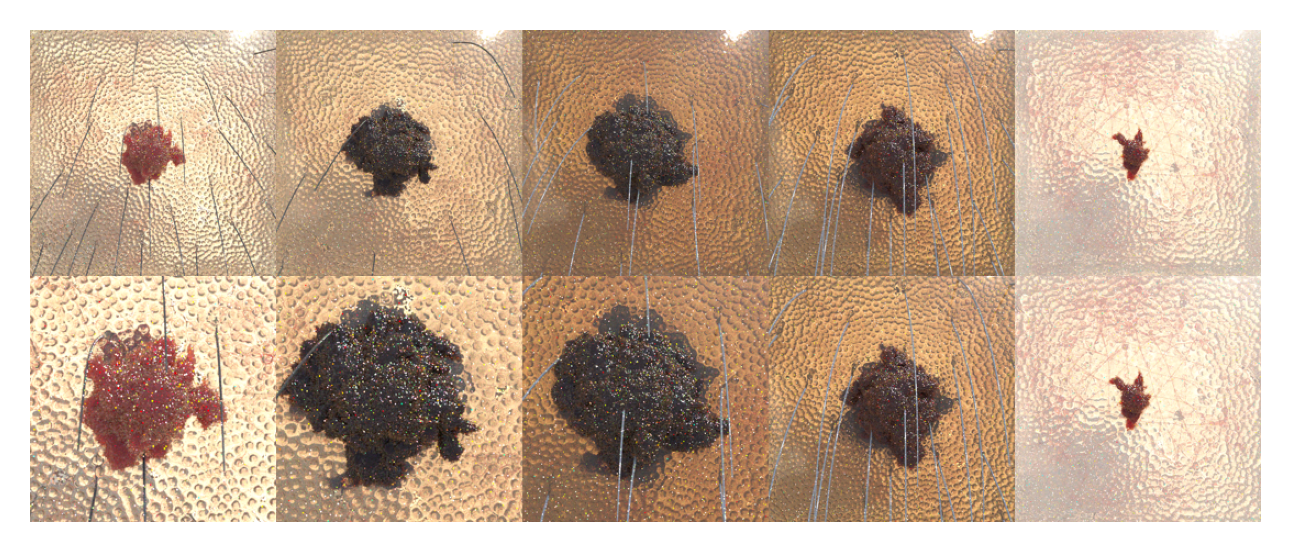}}
    \subfigure[Distribution of lesion relative area and lesion circularity for the real and synthetic images]{\includegraphics[width=0.32\columnwidth]{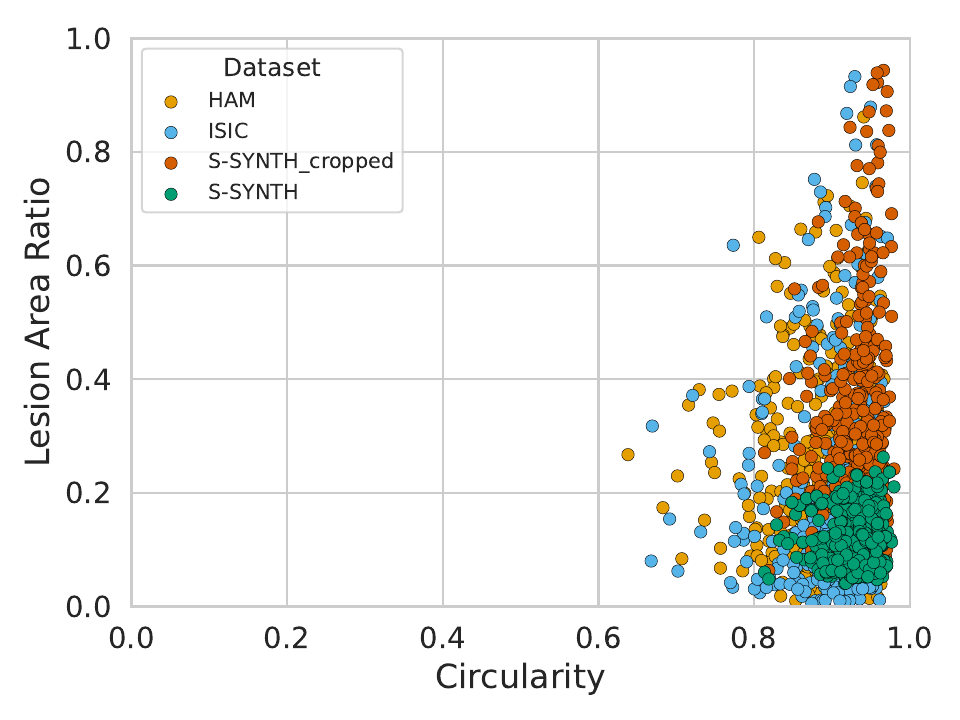}}
    \caption{(a) Visualization of S-SYNTH images before and after random crop of 0-60\% of the original image size around the center (b) Distribution of lesion relative area and lesion circularity for a subset of HAM (500 images), ISIC (259 images), S-SYNTH before cropping (500 images), and S-SYNTH after cropping (500 images).}
    \label{fig:cropped_images}

\end{figure}

 \begin{figure}[h!]
 \centering
 \subfigure[Different proportions of real data]{\includegraphics[width=0.32\columnwidth]{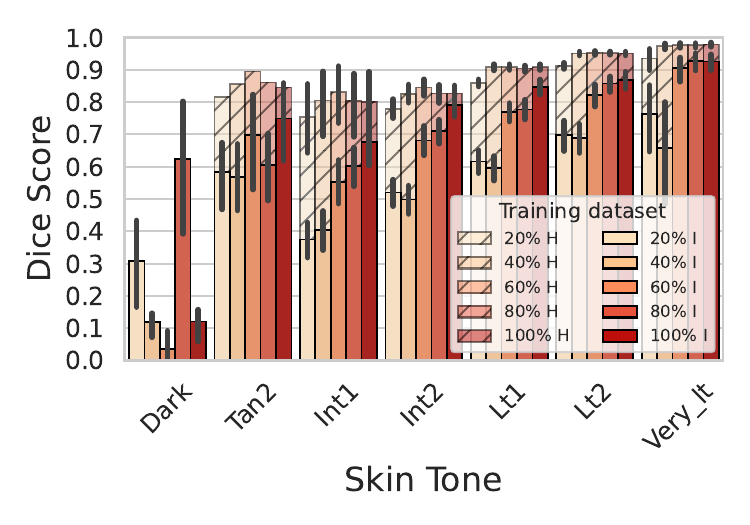}}
 \subfigure[Real data replaced with synthetic data]{\includegraphics[width=0.32\columnwidth]{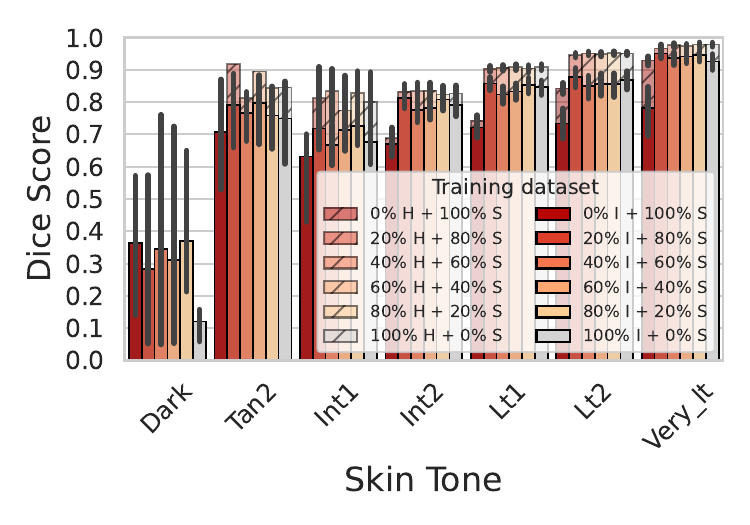}}
  \subfigure[Synthetic data added to real data]{\includegraphics[width=0.32\columnwidth]{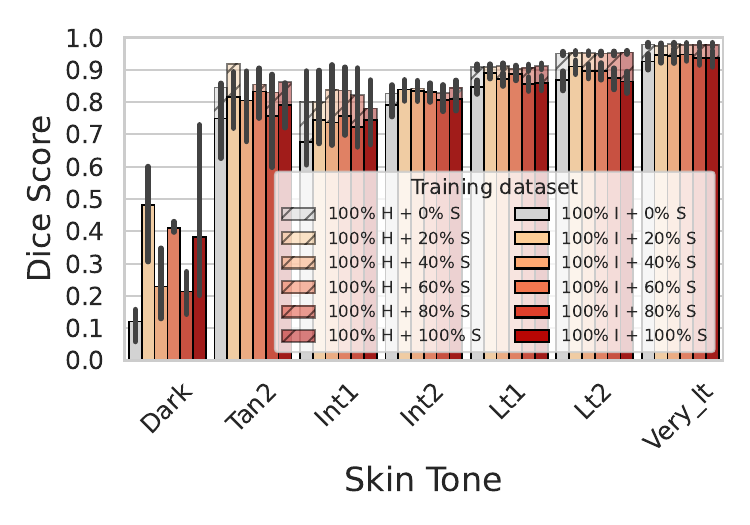}}
 \caption{Changes of model performance when the training data is composed of (a) different subsets of the real images, (b) different subsets of synthetic images substituted for real images (c) different subsets of synthetic images added to real images. The dice scores are stratified based on the skin tone of the test dataset. H: HAM, I: ISIC, S: S-SYNTH.}
 \label{fig:percent_real_stratified}
\end{figure}

\end{document}